\documentclass[journal]{IEEEtran}
\ifCLASSOPTIONcompsoc
  \usepackage[nocompress]{cite}
\else
  \usepackage{cite}
\fi
\ifCLASSINFOpdf
\else
\fi
\usepackage{colortbl}
\usepackage{algorithm}
\usepackage{algorithmicx}
\usepackage{algpseudocode}
\usepackage{lipsum}
\usepackage{amsmath}
\usepackage{combelow}
\usepackage[normalem]{ulem}
\usepackage{amsfonts}
\usepackage{tabu}                      % only used for the table example
\usepackage{booktabs}                  % only used for the table example
\usepackage{xcolor}
\usepackage{soul}
\ifCLASSINFOpdf
  \usepackage[pdftex]{graphicx}
  \DeclareGraphicsExtensions{.pdf,.jpeg,.png}
	\usepackage{epstopdf}
\else
  \usepackage[dvips]{graphicx}
  \DeclareGraphicsExtensions{.eps}
  \DeclareGraphicsExtensions{.tiff}
\fi

\definecolor{inchworm}{rgb}{0.7, 0.93, 0.36}

\usepackage{amsmath}
\usepackage{array}
\newcolumntype{P}[1]{>{\centering\arraybackslash}p{#1}}
\newcolumntype{M}[1]{>{\centering\arraybackslash}m{#1}}

\begin{document}
\title{Detection of Disengagement from Voluntary Quizzes: An Explainable Machine Learning Approach in Higher Distance Education}

\author{Behnam Parsaeifard,
        Christof Imhof,
        Tansu Pancar,
        Ioan-Sorin Comsa,
        Martin Hlosta, \\
        Nicole Bergamin and
        Per Bergamin %~\IEEEmembership{Member,~IEEE}
        % <-this % stops a space
\vspace{-0.05in}
\IEEEcompsocitemizethanks{\IEEEcompsocthanksitem B. Parsaeifard, C. Imhof, T. Pancar, I.-S. Comsa, M. Hlosta, and P. Bergamin are with the Institute for Research in Open-, Distance- and eLearning, Swiss Distance University of Applied Sciences, Brig, CH-3900, Switzerland (e-mail addresses: behnam.parsaeifard@ffhs.ch, christof.imhof@ffhs.ch, tansu.pancar@ffhs.ch, ioan-sorin.comsa@ffhs.ch, martin.hlosta@ffhs.ch, per.bergamin@ffhs.ch). N. Bergamin (e-mail address: nicole.bergamin@ffhs.ch) is with Department of Informatics, Swiss Distance University of Applied Sciences, Brig, CH-3900, Switzerland. 
P. Bergamin is also with the North-West University, Potchefstroom, 2531, South Africa.\protect\\
%E-mail: see http://www.michaelshell.org/contact.html
%\IEEEcompsocthanksitem J. Doe and J. Doe are with Anonymous University.
}
% <-this % stops a space
}

\vspace{-0.05in}
\IEEEtitleabstractindextext{%
\begin{abstract}
Students disengaging from their tasks can have serious long-term consequences, including academic drop-out. This is particularly relevant for students in distance education. One way to measure the level of disengagement in distance education is to observe participation in non-mandatory exercises in different online courses. In this paper, we detect student disengagement in the non-mandatory quizzes of 42 courses in four semesters from a distance-based university. We carefully identified the most informative student log data that could be extracted and processed from Moodle. Then, eight machine learning algorithms were trained and compared to obtain the highest possible prediction accuracy. Using the SHAP method, we developed an explainable machine learning framework that allows practitioners to better understand the decisions of the trained algorithm. The experimental results show a balanced accuracy of 91\%, where about 85\% of disengaged students were correctly detected. On top of the highly predictive performance and explainable framework, we provide a discussion on how to design a timely 
intervention to minimise disengagement from voluntary tasks in online learning.
\end{abstract}

\vspace{-0.05in}
\begin{IEEEkeywords}
% Computer Society, IEEE, IEEEtran, journal, \LaTeX, paper, template.
Behavioural Disengagement, Explainable Machine Learning, Learning Analytics, Distance Education.
%eLearning
\vspace{-0.05in}
\end{IEEEkeywords}}
\maketitle
\IEEEdisplaynontitleabstractindextext
\IEEEpeerreviewmaketitle

\vspace{-0.05in}
\section{Introduction}\label{sec:introduction}
\vspace{-0.05in}
\IEEEPARstart{T}{he} advent of distance education has made learning more flexible than ever before. Instead of having to attend classes and solve tasks at specific time, students are granted more freedom in choosing when to engage with their academic workload. This flexibility attracts many non-traditional student groups to higher education, including students that are employed outside of their studies, either fully or part-time. While deadlines are still set in place, students are responsible themselves for planning and time management, especially as far as non-mandatory tasks and exercises are concerned. This freedom can also lead to satisficing behaviour,  meaning students only do the bare minimum to pass their courses (see e.g., \cite{hadsell2008grade, stohs2016cs}). It can be assumed that such minimalist strategies are quite prevalent due to students being engaged in duties outside their studies (for example work, family and leisure) \cite{marland1990}. The COVID-19 pandemic is thought to have fostered this kind of behaviour even more \cite{schudde2022age}.

Non-completion of voluntary tasks, such as optional quizzes, is a form of behavioural disengagement strongly linked to academic drop-out or attrition \cite{brint2012portrait, harper2009beyond, blondal2012student, schaeffer2010impact}. Voluntary quizzes are critical predictors of later performance (e.g., summative examinations) and play a significant role in academic success \cite{kibble2011voluntary, forster2018feedback, figueroa2021investigating}. Drop-out rates in distance education are notably higher compared to traditional classrooms, particularly among fully-employed students \cite{carr2000distance, stoessel2015sociodemographic}. Early detection of disengagement enables timely interventions and minimise negative impacts. An automated detection system can process students’ interactions with the Learning Management System (LMS) in real-time to predict disengagement behaviours based on log data accumulated throughout the semester. For effective interventions, it is crucial to identify the specific factors contributing to disengagement, such as low participation, inconsistent activities, or irregular access to course materials. These insights enable tailored support, helping students and teachers understand the causes and take corrective actions.

This paper proposes a framework that utilises Machine Learning (ML) to detect student disengagement and employs an explainable approach to interpret the model’s decisions, enabling the design of targeted interventions. Our primary focus is on quiz non-completion as an indicator of disengagement, particularly in voluntary tasks, as completing these tasks is closely linked to improved academic performance. The study sample includes adaptive courses where quiz difficulty is tailored to student performance, chosen for their consistent number of quizzes compared to other courses. To address these objectives, we formulate the following research questions:

\begin{itemize}
    \vspace{-0.02in}
    \item \textbf{RQ1:} What data can be used to detect disengagement from voluntary tasks in online courses? 
    \item  \textbf{RQ2:} Which ML algorithms provide the highest predictive performance in detecting disengagement?
    \item \textbf{RQ3:} How might the explainability of ML decisions be interpreted in the context of potential interventions?
    {\item \textbf{RQ4:} How can 
    %dialog-based 
    intervention strategies be aligned with detected behavioral patterns to inform targeted and timely support actions?}
    \vspace{-0.02in}
\end{itemize}

\noindent To answer these research questions, the following contributions are proposed in this paper:
\begin{itemize}
    \vspace{-0.02in}
    \item Data from four academic semesters at a distance university of applied sciences, encompassing 42 online courses, was collected. In these courses, adaptive tasks with varying difficulty levels were recommended to students based on their performance. (\textbf{RQ1})
    \item Eight ML models are implemented, configured, trained, and evaluated in a comparative analysis, using various performance metrics. The model with the highest predictive accuracy is selected for further investigation to enhance the explainability and interpretation of its predictions. (\textbf{RQ2})
    \item SHapley Additive exPlanations (SHAP) were used to analyze why certain students are at risk of disengaging from voluntary tasks. This analysis highlights key predictors, grouped into three behavioural patterns: erratic, delayed, and irregular. Disengagement predictions are further categorised into low (L), medium (M), and high (H) risk levels, enabling teachers to initiate personalised, dialogue-based interventions tailored to each risk category. (\textbf{RQ3})
    {\item Based on a structured literature review, 
    %dialog-based 
    intervention strategies were identified and mapped to the behavioural patterns derived from SHAP explainability. For each risk level (L, M, H) specific, evidence-based interventions are proposed. Additionally, the study outlines recommended timing for intervention deployment, grounded in the model's detection outputs and the interpretability of behavioral indicators. (\textbf{RQ4})}
    \vspace{-0.02in}
\end{itemize}

\noindent The remainder of the paper is organised as follows: Section \ref{sec:related_works} presents the related work on student disengagement and prediction models, in distance education. Section \ref{sec:methods} outlines the methods and tools used for data collection and processing, along with a description of the proposed explainable ML-based framework. Section \ref{sec:results} presents the findings and addresses the research questions. Section \ref{sec:discussion} offers practical insights for teachers to initiate tailored, personalised dialogue-based interventions. Finally, the paper concludes with Section \ref{sec:conclusion}.

\vspace{-0.05in}
\section{Related Work}\label{sec:related_works}
\subsection{Background on Disengagement in Distance Education}
Engagement is a construct often encountered in education. It is described as “energy in action, the connection between person and activity” (see\cite{russell2005schooling}, p.1) and reflects a person’s active involvement in a task or activity \cite{reeve2004enhancing}. Engagement has been linked to academic performance, persistence, satisfaction, and retention, and is acknowledged to play a significant role in learning, with some researchers going as far as to call it “the holy grail of learning” (see\cite{sinatra2015challenges}, p.1).  Engagement tends to be thought of as behavioural, emotional, and cognitive manifestations of motivation \cite{skinner2009engagement, appleton2008student} and, in contrast to motivation, cannot be separated from its context since an individual is always engaged in something, e.g. a task, an activity or a relationship \cite{fredricks2012measurement}. Currently, no definition of engagement has been universally accepted and there is little to no consistency or specificity in the operationalisation of the construct \cite{halverson2016conceptualizing}. However, some researchers approach the concept of engagement by contrasting it with disengagement, which is characterised by dis-involvement, apathy and a lack of attention \cite{yazzie2007voices}. The same three dimensions as in engagement can also be distinguished here, i.e., cognitive, affective, and behavioural disengagement \cite{trowler2010student}.

Academic disengagement often begins with minor behavioural self-management issues, such as unpreparedness for classes, task incompletion, difficulty with independent work, or procrastination \cite{briesch2009review, Steel2007}. Left unaddressed, it can escalate into chronic absenteeism, poor performance, and ultimately academic drop-out, which can lead to debt, diminished employment prospects, and harm to institutional reputations \cite{archambault2009student, allensworth2013use, chipchase2017conceptualising}. Contributing factors include stress, where disengagement acts as a self-preservation strategy \cite{baik2015first, mann2001alternative}, and a lack of motivation driven by external pressures, involuntary course attendance, or limited alternatives \cite{baik2015first, hunter2012continuum}. Additional causes include unpreparedness for the demands of higher education, unmet expectations (e.g., grades), competing demands outside studies, and institutional challenges, such as low educator-to-student ratios and insufficient support \cite{baldwin2012enhancing, chipchase2017conceptualising}.

\vspace{-0.15in}
\subsection{Prediction Models used in Behavioural Disengagement}
Given the potentially severe consequences of academic disengagement, intervention strategies have been devised to counteract this behaviour, which rely on the early detection of disengagement. The focus should be set on objective, non-intrusive methods of measurement (i.e., log data analysis). These methods are far more suitable than questionnaires or other subjective measures when trying to detect disengagement during the academic semester. Objective indicators of academic disengagement include participation (e.g., in forum discussions), time spent studying, effort spent studying (indicated e.g., by using support services or reviewing notes), accessing the LMS, the completion of assessments, academic performance, and collaborative efforts \cite{chipchase2017conceptualising}.

The study by \cite{cocea_2009} predicts disengagement in eLearning systems like HTML-Tutor and iHelp using statistical indicators (mean, standard deviation) from log data on activities such as reading pages, pre-tests, assessments, and feedback pages. In \cite{hussain_2018}, various ML models are compared for engagement detection based on final grades, using predictors such as clicks on forums and assessments. Similarly, the approach in \cite{raj_2022} utilises the Open University Learning Analytics Dataset (OULAD) to classify engagement based on metrics like clicks and exam scores. Behavioural disengagement prediction is explored in \cite{mills_2014} in an online environment focused on reading tasks, where models predict quitting behaviour at different stages using time-spent features. These studies place limited emphasis on model explainability and the development of intervention strategies. In a related approach \cite{saqr_trajectory_2021},  clustering methods are used to identify engagement patterns from data features like activity frequency and regularity, resulting in three engagement states: active, average, and disengaged. Using Hidden Markov Models, they trace these engagement trajectories across an educational program, finding early dropouts within the disengaged group, underscoring the importance of early interventions.

Building on the state-of-the-art approaches discussed, this paper proposes an ML framework to detect student disengagement in voluntary and adaptive online exercises. The analysed key features include online inactivity, preferred learning times, and irregularity in accessing and completing quizzes. In addition to achieving high accuracy in identifying disengaged students, the framework emphasises the explainability and interpretability of machine learning.

\vspace{-0.15in}
\section{Methods}\label{sec:methods}
The following aspects will be covered in this section: \textit{a)} courses and participants used in data collection, \textit{b)} data collection, preprocessing and feature selection; \textit{c)} machine learning modeling, training and optimisation; \textit{d)} strategies to explain and interpret the ML decisions to support teachers or practitioners in designing tailored interventions.

\begin{figure}[t]
    \centering
    \includegraphics[width=0.8\columnwidth]{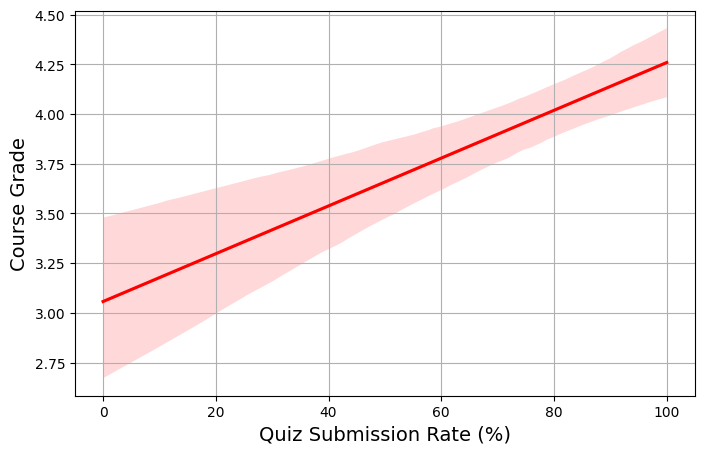}
    \vspace{-0.1in}
    \caption{Relationship between quiz submission rate (engagement) and final course grade (performance) ranging from 1 (minimum) to 6 (maximum). }
    \label{fig:engagement_vs_performance}
    \vspace{-0.2in}
\end{figure}

\begin{figure}[t]
    \centering
    \includegraphics[width=0.8\columnwidth]{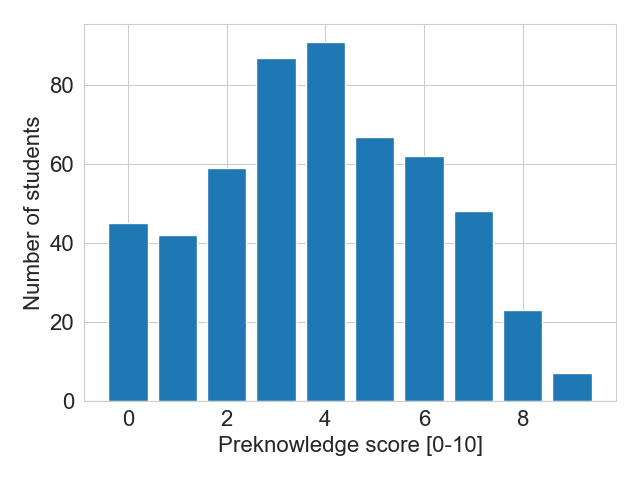}
    \vspace{-0.15in}
    \caption{Histogram of the number of students against preknowledge test scores. }
    \label{figure_1}
    \vspace{-0.25in}
\end{figure}

\vspace{-0.15in}
\subsection{Courses and Participants} 
This study was implemented in a distance learning university located in Central Europe, integrating a sophisticated adaptive learning framework within a blended learning model. The educational approach combines an 80\% online component with 20\% face-to-face interaction, utilizing Moodle as the backbone for delivering adaptive courses. These courses are uniquely structured around quiz activities, serving as both the instrument for student engagement and the mechanism for academic assessment \cite{censored_ref_1, censored_ref_2}.

{Building on this educational setup, we analyzed data from the adaptive courses to examine how voluntary quiz submission (used as a proxy for student engagement) relates to academic performance. As shown in Figure \ref{fig:engagement_vs_performance}, there is a clear upward trend between the rate of non-mandatory quiz submissions and final course grades, which are measured on a scale from 1 (lowest) to 6 (highest). This relationship confirms that students who engage more consistently with voluntary learning activities tend to achieve better academic outcomes. In this context, detecting disengagement is essential for identifying students who are beginning to withdraw from optional course components. Recognizing these patterns early allows instructors to respond with timely, targeted support before disengagement further affects academic performance.}

\begin{figure*}[t]
    \centering
    \includegraphics[width=0.9\textwidth]{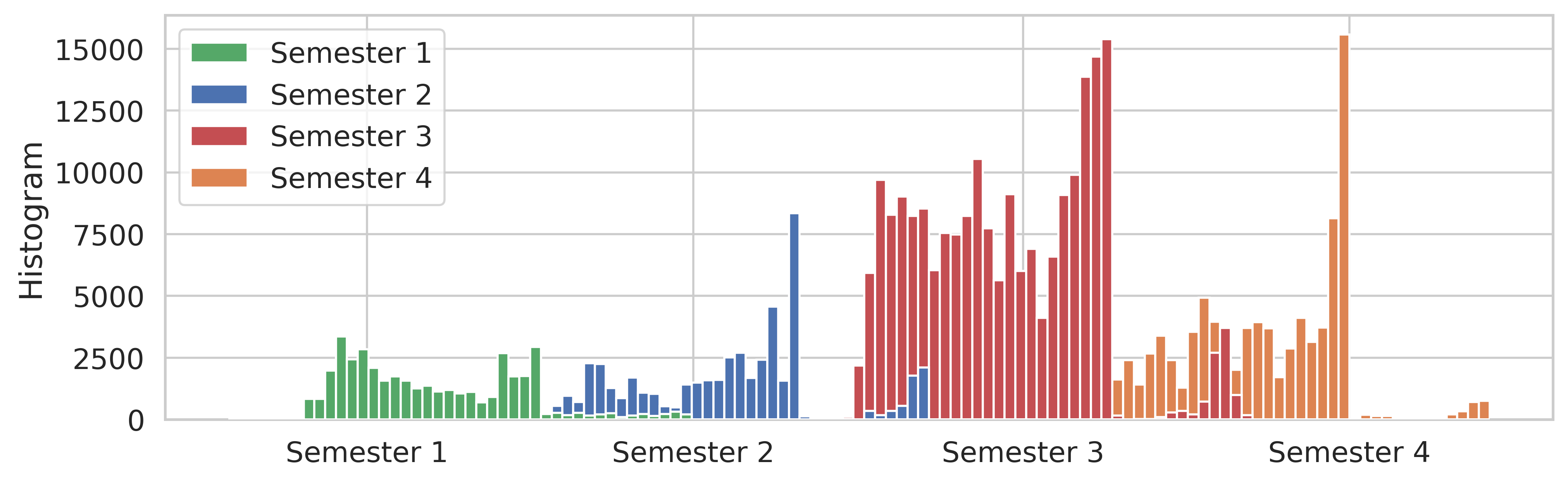}
    \vspace{-0.05in}
    \caption{Histogram of students' activities (view, review, etc.) in quizzes in each semester.}
    \label{figure_2}
    \vspace{-0.2in}
\end{figure*}

The onset of the adaptive learning experience involves an assessment of prior knowledge, which identifies students as either novices, with less than 50\% score on preliminary exercises, or as experts, with score of 50\% or higher. The histogram of the preknowledge test scores is shown in Figure~\ref{figure_1}. This classification informs the level of detail and support provided in subsequent quiz activities. Novices receive quizzes with detailed instructions, ensuring adequate support, while experts are challenged with quizzes that encourage independent problem-solving \cite{censored_ref_1,  censored_ref_2}.

Each course is organised into discrete blocks (5 blocks per course, in general), with quizzes at the beginning of each block designed to match the learner’s knowledge state determined by the prior knowledge assessment. The adaptive system recommends future quizzes based on performance, creating a personalised learning path. Feedback is tailored and provided through Moodle, which adapts to the students' responses, offering specific guidance to rectify misunderstandings and promote understanding \cite{ censored_ref_2, durlach2011designing}. Furthermore, the adaptive design fosters the development of problem-solving and transfer skills, enabling students to apply their knowledge to new and varied situations \cite{van2002effect}.

This facet of the adaptivity mechanism is critical in preparing students to synthesise and adapt their learning to real-world contexts.
The scope of this analysis encompasses data from five topics: Introductory Mathematics, Mathematics 1, Mathematics 2, Probability and Statistics, and Applied Natural Sciences II, spanning four semesters: fall semester 2017/2018 (semester 1), spring semester 2018 (semester 2), fall semester 2018/2019 (semester 3), and spring semester 2019 (semester 4). Figure~\ref{figure_2} presents histograms of the number of interactions within voluntary quizzes, broken down by week and by semester. A higher volume of interactions is observed towards the end of each semester, indicating that students are more motivated to complete exercises as a preparation strategy for final exams. In semesters 3 and 4, the number of courses increases significantly compared to earlier semesters, which explains the higher interaction count. Additionally, some activities from previous semesters are repeated in subsequent semesters for students who received unsatisfactory grades and are participating in reexaminations. Over all four semesters, an aggregate number of 565 students' quiz attempts and the corresponding Moodle logs were meticulously examined. Notably, the system permits multiple quiz attempts, with detailed logs of each attempt captured within the database. 
%\textcolor{blue}{In Section \ref{sec:introduction}, we defined non-mandatory quiz completion as engagement and mentioned that it is positively linked to academic performance. To show this, we show the relationship between the quiz submission rate and the final course grade, as indicator of the performance, in Figure \ref{fig:engagement_vs_performance}. This figure illustrates a positive association between students' engagement, measured by the percentage of voluntary quizzes submitted, and their final course performance. Given this relationship, predicting engagement through machine learning---especially in early stages of the course---could enable timely interventions. Identifying students with low predicted engagement may allow instructors to provide targeted support, thereby potentially improving academic outcomes.}

\vspace{-0.1in}
\subsection{Data Collection}\label{sec:data_collection_and_preprocessing}
In Moodle, data is organised into multiple related tables, explained below. The two primary data sources for our study are the quiz and log tables. The quiz table includes details about the courses associated to quizzes, the students attempting the quizzes, and the status of each quiz (i.e., whether it has been submitted or is still in progress), etc. Key columns from the quiz table are described in Table~\ref{tab:quiz_table}. In contrast to the quiz table, which provides final quiz results for each student (including submissions and grades), the logs table offers detailed information about student interactions with Moodle. Every interaction is defined by three elements: a \textit{component}, an \textit{event}, and an \textit{objectID}. When a student attempts a quiz, the \textit{objectID} matches the \textit{attemptID} in the quiz table (Table~\ref{tab:quiz_table}), the component is labeled as \textit{"quiz"} and the event can be, for instance, \textit{"view"} or \textit{"submit"}, indicating whether the student viewed or submitted the quiz during that interaction. Key columns from the log table are described in Table~\ref{tab:log_table}.

\begin{table}[t]
  \caption{Description of the quiz table.}
  \label{tab:quiz_table}
  \centering
  \begin{tabular}{r l}
    \toprule
    \multicolumn{1}{c}{\textbf{Column}} & \multicolumn{1}{c} {\textbf{Description}} \\
    \midrule
    \textit{quizID} & Unique identifier for the quiz \\
    \textit{courseID} & Unique identifier for the corresponding course \\
    \textit{attemptID} & Unique identifier for the student's attempt \\
    \textit{attemptNo} & Student's current attempt number \\
    % \midrule 
    \textit{startTime} & Start time of the attempt \\
    \textit{endTime} & End time of the attempt (in case of submission) \\
    \textit{quizStatus} & Status of the quiz (in-progress/submitted) \\
    \textit{maxPoints} & Maximum number of points of the quiz \\
    \textit{points} & Number of points achieved by student (if submitted)\\
    \bottomrule
  \end{tabular}
  \vspace{-0.15in}
\end{table}

\begin{table}[t]
  \caption{Description of the logs table.}
  \label{tab:log_table}
  \centering
  \begin{tabular}{r l}
    \toprule
    \multicolumn{1}{c}{\textbf{Column}} & \multicolumn{1}{c} {\textbf{Description}} \\
    \midrule
    \textit{studentID} & Unique identifier for the student \\
    \textit{courseID} & Same as courseID in Table \ref{tab:quiz_table} \\
    \textit{objectID} & Same as attemptID in Table \ref{tab:quiz_table} if the component is 'quiz' \\
    \textit{component} & Type of student's interaction with LMS (quiz, module)\\
    \textit{event} & Action taken by the student (e.g., view, submission, etc.) \\
    \textit{timestamp} & Timestamp of the action \\
    \textit{origin} & Platform used by the student (e.g., web or mobile) \\
    \bottomrule
    \vspace{-0.15in}
  \end{tabular}
  \vspace{-0.2in}
\end{table}

It's important to note that the logs table contains a significantly larger number of rows compared to the quiz table. By joining these two tables using the common column, namely \textit{attemptID} (which is unique for each student attempt across all quizzes and courses), we obtain a consolidated table with columns such as \textit{courseID}, \textit{quizID}, \textit{studentID}, \textit{attemptID}, \textit{component}, \textit{event}, and \textit{timestamp} representing student interactions with the LMS during that attempt.

\begin{table*}[t]
\centering
\caption{An example of student activity data.}
\label{tab:student_activity_2}
\begin{tabular}{|p{1cm}|p{2cm}|l|p{1cm}|p{1.25cm}|p{1cm}|p{5cm}|p{0.8cm}|}
\toprule
attempt id & activity on day & date\_rel & inactive days & submission day & days before submit & accumulated activity & did submit  \\
\midrule
$x$ & $[E_{start}, E_1, E_2]$ & $0$ & $0$ & $6$ & $6$ & $[E_{start}, E_1, E_2]$ & $N$  \\
$x$ & $[ ]$ & $1$ & $0$ & $6$ & $5$ & $[E_{start}, E_1, E_2]$ & $N$  \\
$x$ & $[E_3]$ & $2$  & $1$ & $6$ & $4$ & $[E_{start}, E_1, E_2, E_3]$ & $N$ \\
$x$ & $[ ]$ & $3$ & $1$ & $6$ & $3$ & $[E_{start}, E_1, E_2, E_3]$ & $N$  \\
$x$ & $[E_4, E_5]$ & $4$ & $2$ & $6$ & $2$ & $[E_{start}, E_1, E_2, E_3, E_4, E_5]$ & $N$  \\
$x$ & $[ ]$ & $5$ & $2$ & $6$ & $1$ & $[E_{start}, E_1, E_2, E_3, E_4, E_5]$ & $N$ \\
$x$ & $[E_6, E_{end}]$ & $6$ & $3$ & $6$ & $0$ & $[E_{start}, E_1, E_2, E_3, E_4, E_5, E_6, E_{end}]$ & $Y$ \\
\bottomrule
\end{tabular}
\vspace{-0.2in}
\end{table*}

\vspace{-0.15in}
\subsection{Data Preprocessing}\label{sec:preprocessing}
The collected data contains a record for each interaction of the student in an attempt. We initially preprocess the raw data and gather records on a daily basis. Then, we calculate for each day the following new features:
\begin{itemize}
    \vspace{-0.05in}
    \item \textbf{date\_rel}: day number relative to the start of the attempt.
    \item \textbf{activity on day}: activities performed by the student on that particular day; each activity can be one of the possible events described in Table~\ref{tab:log_table}.
    % \item \textbf{inactive days since last activity}: number of days since the last activity
    \item \textbf{inactive days}: total number of inactive days since the start of the attempt
    \item \textbf{submission day}: day on which the student submitted their attempt.
    \item \textbf{days before submit}: days elapsed between the activity of that day and submission (if applicable).
    \item \textbf{accumulated activity}: list of all activities from the start of the attempt up to and including the current day.
    \item \textbf{did submit}: indicates whether the student submitted the attempt on that day ('$Y$' for yes, '$N$' for no).
    \vspace{-0.00in}
\end{itemize}

An example of preprocessed student data for an attempt is shown in Table~\ref{tab:student_activity_2}. The table shows that, on day $0$, the student initiated the attempt, $E_{start}$ and engaged in activities $[E_1, E_2]$, where $E_i$ contains the activity type (e.g., \textit{view}) as well as the timestamp. By day $2$, following a day of inactivity, they resumed with activity $[E_3]$, contributing to the ongoing accumulation of activities $[E_{start}, E_1, E_2, E_3]$. Moving to day $4$, after another inactive day, the student performed activities $[E_4, E_5]$, further expanding the accumulated activities to $[E_{start}, E_1, E_2, E_3, E_4, E_5]$. On day 6, after yet another day of inactivity, the student performed another activity $[E_6]$ and submitted their attempt ($E_{end}$). The accumulated activities up to this point encompassed $[E_{start}, E_1, E_2, E_3, E_4, E_5, E_6]$ (submission event excluded if it exists). This table structure provides a clear overview of the student's activity progression throughout the attempt, including periods of activity and inactivity, leading up to the submission of the quiz attempt.

Furthermore, we fill in the missing days (days on which the student is inactive) in this table by assigning an empty list to \textit{activity on day}. However, we update other columns such as \textit{date\_rel}, \textit{inactive days}, etc., for these missing days. This approach ensures that the accumulated activity for missing days matches the accumulated activity of the last active day, while other columns are updated accordingly. This method contributes to \textit{1)} data augmentation, and \textit{2)} model robustness, as the model is expected to predict submission at any time. 

\vspace{-0.1in}
\subsection{Featurisation and Problem Formulation}
Given a sequence $\{E_{start}, E_1, E_2, \dots, E_n\}$ of accumulated events (submission event excluded) and the corresponding timestamps $\{T_{start}, T_1, T_2, \dots, T_n\}$ representing a student's interactions during an attempt up to a specific day $t$, we first compute the time differences between each consecutive timestamp, resulting in~$\{\Delta T_1, \Delta T_2, \dots, \Delta T_{n}\}$, where~$\Delta T_1 = T_{1} - T_{start}$, $\Delta T_2=T_2-T_1$, and so on.  Various statistics derived from this sequence, such as min, max, median, mean, standard deviation, skewness, and kurtosis are calculated. These statistics capture the temporal information in the data and make up a part of feature vector. Other features include: attempt number, number of previous attempts, previous performance of the student, inactive days, count of activities in workday/weekend and morning/afternoon/evening. 

By processing the feature vector in each record as per Table~\ref{tab:student_activity_2}, the goal is to predict whether or not the student is engaged. A student is considered as engaged in the voluntary quiz if they submitted the quiz on that day and disengaged otherwise (either submitting after time $t$ or never submitting). Note that in this method, both the feature vector and the label (engaged/disengaged) are time-dependent, meaning they are defined relative to the particular day.

\vspace{-0.15in}
\subsection{ML for Detecting Behavioral Disengagement}
Multiple ML models are implemented in this study, among which, we consider the Random Forest (RF) as our baseline model in detecting student disengagement from voluntary quizzes, since this approach has been shown to be effective for educational predictions \cite{yu2021academic, censored_ref_3, santamaria2020learning}. RF is an ensemble learning method that builds multiple decision trees using random subsets of data and features, and then combines their predictions for improved accuracy of predictions. To address RQ2, we need to test the hypothesis that the RF model outperforms other ML approaches in detecting disengagement. To do this, we implement, configure, train, and evaluate several additional ML models, comparing their performance to determine the best-performing one. The ML models analyzed in this framework are: Decision Tree (DT) \cite{breiman2017classification}, Support Vector Machine (SVM) \cite{cortes1995support}, feed-forward Neural Networks (NN) \cite{rosenblatt1958perceptron}, K-Nearest Neighbors (KNN) \cite{cover1967nearest}, Logistic Regression (LR) \cite{cox1958regression}, Naïve Bayes (NB) \cite{maron1960relevance},  XGBoost (XGB) \cite{xgb_2016}, and Gradient Boosting Machines (GBM) \cite{friedman2001greedy}.

In this study, we use a semester-based data split approach: specifically, data from the first three semesters (semesters 1, 2, and 3 in Fig.~\ref{figure_2}) is used as training data to configure and train the models, while data from the final semester (semester 4 in Fig.~\ref{figure_2}) serves as test data for performance evaluation. This approach assesses the models' ability to generalise across academic terms and simulates real-world scenarios. 

The objective of the ML model is to accurately identify disengaged students in new, unseen data for timely interventions while minimizing unnecessary interventions for engaged students. Achieving optimal performance requires configuring several hyperparameters specific to the model being used. For instance, in the case of the RF model, key hyperparameters such as the number of decision trees and the maximum depth of each tree need to be set. Similarly, for a neural network, determining the optimal number of hidden layers and nodes is crucial for effective learning. To identify the best configuration and ensure the model generalizes well, a 4-fold cross-validation is employed to maximize the Area Under the Curve (AUC) of the True Positive Rate (TPR) and False Positive Rate (FPR). Once the optimal hyperparameters are determined, the models are trained on the full train dataset and evaluated using the test dataset. Table~\ref{tab:metrics} provides an overview of the evaluation metrics used to assess and report the performance of the proposed ML models.
\begin{table*}[t]
\centering
\caption{Metrics used for model evaluation}
\label{tab:metrics}
\renewcommand{\arraystretch}{1.3} % Adjust row spacing
\begin{tabular}{m{4cm} m{2.75cm} m{10cm}}
\toprule
\textbf{Metric} & \textbf{Definition} & \textbf{Description} \\ 
\midrule
True Positive (TP) & & Number of engaged students that are correctly detected (as engaged)  \\ 
False Positive (FP) & & Number of disengaged students that are incorrectly detected (as engaged) \\ 
True Negative (TN) & & Number of disengaged students that are correctly detected (as disengaged)  \\ 
False Negative (FN) & & Number of engaged students that are incorrectly detected (as disengaged) \\ 
Positive Class (P) & $TP + FN$ &  Number of engaged students \\ 
Negative Class (N) & $TN + FP$ & Number of disengaged students \\ 
Positive Predictive Value (PPV) (precision) & $\frac{TP}{TP+FP}$ & Proportion of detected engaged students that are correct. \\ 
Negative Predictive Value (NPV)  & $\frac{TN}{TN+FN}$&  Proportion of detected disengaged students that are correct.  \\ 
True Positive Rate (TPR) (recall/sensitivity) & $\frac{TP}{TP+FN}$ &  Proportion of engaged students that are correctly detected. \\ 
True Negative Rate (TNR) & $\frac{TN}{TN+FP}$ & Proportion of disengaged students that are correctly detected. \\ 
False Positive Rate (FPR) & $\frac{FP}{TN+FP}=1-TNR$ & Proportion of disengaged students that are incorrectly detected.  \\ 
False Negative Rate (FNR)  & $\frac{FN}{TP+FN}=1-TPR$ & Proportion of engaged students that are incorrectly detected. \\ 
F1 Score (engaged class) & $2\frac{PPV\times TPR}{PPV+TPR}$& Measures model’s ability to correctly identify engaged students. \\ 
F1 Score (disengaged class) & $2\frac{NPV\times TNR}{NPV+TNR}$ & Measures model’s ability to correctly identify disengaged students. \\ 
Balanced Accuracy (BA) & $\frac{TPR+TNR}{2}$& Measures model’s ability to correctly classify both engaged and disengaged students while accounting for class imbalance.  \\ 
Area Under the Curve (AUC) & & Measures model's ability to distinguish between the engaged and disengaged students. \\ 
\bottomrule
\end{tabular}
\vspace{-0.1in}
\end{table*}

\vspace{-0.15in}
\subsection{Explanation and Interpretation of ML Predictions}\label{sec:expl_interpret}
In ML applications involving human-centric data, such as education, the significance of explainability cannot be overstated. Beyond achieving accurate predictions, models must offer transparency and interpretability. This is particularly critical in educational contexts, where teachers require a clear understanding of model decisions to make informed interventions. Numerous studies emphasise the importance of explainability in machine learning within the educational domain \cite{drachsler2016privacy, khosravi2022explainable, holmes2022ethics}. The General Data Protection Regulation (GDPR) Article 13 specifically identifies the need for understanding the logic behind automated decision-making processes\cite{webb2021machine}.

Models like LR and SVM with linear kernels are easier to interpret because their decisions are based on assigning coefficients to each feature. However, despite their simplicity, linear models generally perform poorly on educational datasets, which are often not linearly separable. While non-linear, decision trees are also interpretable, as they provide a clear decision-making path from the root to the leaves. However, more complex models (e.g., RF, GBM, XGB, NN) have a greater capacity to capture non-linear relationships. Yet, these models are more difficult to interpret due to their complexity. Therefore, explainable ML algorithms are needed to provide insights into how these models make decisions. These techniques help to break down the "black box" nature of complex models, enabling teachers to understand the contribution of different features in models' decisions and ensure that the models are making fair and reliable decisions.

In this paper, we use the SHAP method \cite{NIPS2017_7062} to explain the predictions of machine learning models. SHAP is a model-agnostic approach, inspired by cooperative game theory, that assigns each feature's contribution to the final prediction, similar to calculating a player’s contribution in a game. Based on SHAP values, we define three kinds of behaviours:
\begin{itemize}
\item \textbf{Erratic Behavior}: reflects a lack of consistent activities in the student’s approach to completing voluntary tasks and difficulties in effective time management.

\item \textbf{Delayed Behaviour}: suggests that the student initially started the task but then experienced significant delays in further progress, which can result from disengagement, poor time management, or a lack of motivation.

\item \textbf{Irregular Behaviour}: indicates that the student is exploring the quiz but is hesitant or uncertain about completing it. This may point to a lack of confidence, difficulty in understanding the material, or potential disengagement. 
\end{itemize}

These patterns, uncovered through SHAP explanations, help us categorise and interpret the risks of behavioral disengagement based on our collected data, including:
\begin{itemize}
\item \textbf{High Risk}: occurs when both erratic and delayed behaviours are detected in a student's data.
\item \textbf{Medium Risk}: assigned when either erratic behaviour or delayed behaviour is detected in student's data.
\item \textbf{Low Risk}: given when only the irregular behavioural pattern is identified in student's data.
\end{itemize}

Once the risk level is determined, targeted interventions can be designed to engage students in voluntary tasks. In cases of high risk, interventions should prioritise addressing erratic behavior first, followed by strategies to tackle delayed engagement. When a medium risk of disengagement is detected, then the intervention should be designed to mitigate the erratic or the delayed behaviour. More information on how to design and implement these interventions are provided in Section \ref{sec:designing-dialog}.

\vspace{-0.1in}
\section{Results}\label{sec:results}
{The aim of this section is to address the four research questions presented in Section~\ref{sec:introduction}. We begin by examining the dataset used to train the machine learning models. This is followed by a comparative evaluation of multiple models to identify the one with the highest performance for further analysis. We then analyze how specific data features contribute to model predictions and how these relate to identifiable behavioral patterns. Building on the interpretability provided by SHAP, we integrate insights from the literature to identify relevant 
%dialog-based 
intervention strategies. These strategies are then adapted to the context of our framework for detecting disengagement from voluntary quizzes in adaptive online courses, with consideration given to the timing and targeting of interventions based on risk levels.}
%The aim of this section is to address the three research questions outlined in Section \ref{sec:introduction}. First, we will explore the data to be used to train the ML models. Next, based on a comparative analysis, we will select the best performing ML method for further analysis. We will then discuss how specific data features relate to certain behavioral patterns and outline how model predictions should be interpreted. Based on the interpretion of ML predictions, we indentify from literature the support of dialog based interventions and we tailor this for the applicability in our scenario when detecting dissengagement from voluntary quizzes in adaptive courses.

\vspace{-0.1in}
\subsection{RQ1: What data can be used to detect disengagement from voluntary tasks in online courses?}
\vspace{-0.05in}
We consider three categories of data features: attempts-related, performance-related, and activity-related features. Attempts-related features include the number of attempts and prior attempts in other quizzes. Performance-related features are derived from previous performance of the student in prior attempts. Activity-related features are divided into period-related, inactivity-related, and temporal statistics. Period-related features track activity counts across weekdays and weekends, split by time of day (morning, afternoon, evening). Inactivity-related features measure the total inactive days since the quiz began. Temporal statistics are based on the differences between consecutive activities, and include descriptive metrics, such as mean, median, standard deviation, kurtosis, and skewness.

These statistics provide critical insights into student behavior and engagement patterns. For instance, the minimum value (\textit{stat\_min}) shows how quickly students engage after the quiz starts. The mean (\textit{stat\_mean}) shows average time difference between consecutive activities — lower values suggest regular engagement, while higher values may indicate sporadic participation or disengagement. The standard deviation (\textit{stat\_sd}) measures the spread of activity timing, with a low value indicating consistent engagement and a high value pointing to disengagement behavior.

Table~\ref{tab:mean-sd-features} provides a comparison of mean values and standard deviations for the features introduced above categorised by submission status (did submit or not) for the train data (semesters 1, 2 and 3 from Fig.~\ref{figure_2}). For features related to activity counts, such as \textit{workday\_morning\_count}, those who submitted (Y) generally show higher mean values compared to those who did not submit (N). This trend continues across counts of different times of day (morning, afternoon, evening) and counts on workdays versus weekends (workday, weekend). In terms of \textit{days\_inactive}, those who submitted (Y) have notably lower mean values, indicating more frequent activities. Features like \textit{previous\_attempts} and \textit{previous\_perf} exhibit higher means for submissions (Y), implying more extensive prior engagement and potentially better previous performance.

\begin{table}[t]
\caption{Mean and standard deviation of features grouped by submission status (did submit or not). Unit-based stats (min, median, mean, and sd) are converted from seconds to hours.}
\label{tab:mean-sd-features}
\centering
\begin{tabular}{lll}
\toprule
did\_submit & N & Y \\
\midrule
number of samples & 17,878 & 15,397 \\
workday\_morning\_count & 0.94 ± 3.3 & 2.75 ± 6.57 \\
workday\_afternoon\_count & 2.64 ± 5.05 & 5.25 ± 8.44 \\
workday\_evening\_count & 1.69 ± 4.43 & 3.85 ± 7.61 \\
weekend\_morning\_count & 0.47 ± 2.19 & 1.36 ± 4.94 \\
weekend\_afternoon\_count & 1.06 ± 3.52 & 1.85 ± 5.64 \\
weekend\_evening\_count & 0.26 ± 1.63 & 0.62 ± 3.28 \\
days\_inactive & 4.99 ± 5.47 & 0.28 ± 1.67 \\
attemptnr & 1.18 ± 0.53 & 1.3 ± 0.6 \\
previous\_attempts & 23.06 ± 22.84 & 34.03 ± 29.79 \\
previous\_perf & 0.73 ± 0.29 & 0.89 ± 0.16 \\
stat\_min & 0.0 ± 0.13 & 0.0 ± 0.0 \\
stat\_mean & 0.48 ± 1.54 & 0.12 ± 0.38 \\
stat\_median & 0.21 ± 1.32 & 0.01 ± 0.03 \\
stat\_sd & 0.81 ± 2.11 & 0.36 ± 1.13 \\
stat\_skew & 0.91 ± 1.23 & 1.94 ± 1.25 \\
stat\_kurtosis & 1.32 ± 4.58 & 4.08 ± 6.63 \\
\bottomrule
\end{tabular}
\vspace{-0.2in}
\end{table}

For statistics-based features, the mean and median (\textit{stat\_mean} and \textit{stat\_median}) are notably lower for those who submitted, reflecting quicker completion times. Submitting students also show lower variability (\textit{stat\_sd}), indicating consistent engagement, while non-submitting students display greater unpredictability. Both groups have positive skewness, showing that most activity occurs shortly after the quiz starts; however, higher skewness among the submitting group suggests some students complete quizzes later. Kurtosis values are significantly higher for the submitting group, implying late but consistent engagement without exhibiting delayed behaviour.

\begin{figure}[t]
    \vspace{-0.05in}
    \centering
    \includegraphics[width=8.5cm]{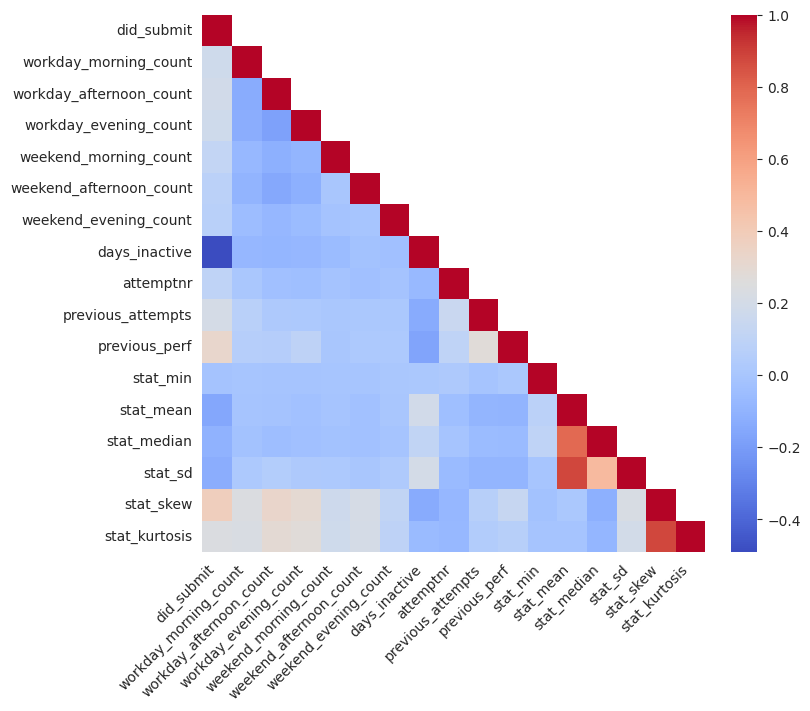}
    \caption{Correlation between features (as well as the target "did\_submit").}
    \label{figure_3}
    \vspace{-0.2in}
\end{figure}

To understand the nature of the data better, we show the Pearson correlation \cite{pearson1895vii} in Figure \ref{figure_3}. A high positive correlation is observed between \textit{stat\_mean}, \textit{stat\_median}, and \textit{stat\_sd} which implies that with an increase in one of these features, the other two features also increase. A relatively high negative correlation is also observed between \textit{did\_submit} and \textit{inactive\_days} implying that as \textit{days\_inactive} increases, the \textit{did\_submit} decreases (changes from submission to non-submission). To answer RQ1, the features used to predict disengagement from voluntary quizzes are those from Table~\ref{tab:mean-sd-features}.

As outlined in Section \ref{sec:expl_interpret}, once a quiz is initiated, it is crucial to categorise the student's interaction with the voluntary quiz based on their behavior. By doing so, the risk levels of disengagement from voluntary quizzes can be assessed, making it possible to develop personalised interventions. These behaviors are classified into four categories: erratic behaviour, delayed behaviour, irregular behaviour, and full engagement. In addressing RQ1, we aim to identify the data features associated with these types of behaviors. Erratic behaviour defined in Section \ref{sec:expl_interpret} is associated with time management (e.g., activity counts during weekdays, and weekend mornings, afternoons, and evenings). Delayed behaviour is primarily indicated by the number of inactive days, while the irregular behaviour suggests tentative engagement, observed through statistics-based features. Full engagement is assigned if none of the above types are detected. The classification of each data sample into one of these categories will depend on the ML algorithm used (discussed in RQ2) and the SHAP explainability (addressed in RQ3).

\vspace{-0.1in}
\subsection{RQ2: Which ML algorithms provide the highest performance in detecting disengagement?}
Four levels of learning analytics are considered: descriptive, diagnostic, predictive, and prescriptive. Descriptive analytics summarises historical data to identify patterns, such as student disengagement. Diagnostic analytics investigates the reasons behind observed patterns. Predictive analytics forecasts future disengagement based on trends and historical data, while prescriptive analytics recommends interventions to mitigate predicted disengagement. This study aligns with diagnostic analytics, leveraging explainable ML models to identify patterns of disengagement and understand the reasons behind them.

This section focuses on developing an accurate and efficient ML model to detect student disengagement in online learning using Moodle interaction data. Eight ML models are implemented using Python's Scikit-learn library \cite{scikit-learn}, and configured using the training data from semesters 1, 2, and 3 (as shown in Fig. \ref{figure_2}). The model configuration was conducted via grid search optimisation of hyperparamters to maximise the AUC performance, yielding 200 trees for RF, a depth of 8 for DT, 1000 trees for XGB, 100 estimators for GBM, and two hidden layers of 100 nodes each for NN. These configurations were used to retrain the models on training data and to evaluate the model on the test dataset from semester 4 (Fig. \ref{figure_2}) to compare model effectiveness and identify the most suitable approach for detecting disengagement.

\begin{table}[t]
\caption{Evaluation metrics for test dataset.}
\label{tab:evaluation-metrics}
\centering
\begin{tabular}{lrrrrrrrr}
\toprule
 & PPV & NPV & TPR & TNR & F1$_{tp}$ & F1$_{tn}$ & BA & AUC \\
\midrule
RF & 0.89 & 0.96 & 0.98 & 0.83 & 0.93 & 0.89 & 0.90 & 0.96 \\
DT & 0.88 & 0.93 & 0.96 & 0.83 & 0.92 & 0.88 & 0.89 & 0.94 \\
XGB & 0.86 & 0.99 & 0.99 & 0.78 & 0.92 & 0.87 & 0.89 & 0.96 \\
GBM & 0.90 & 0.95 & 0.97 & 0.85 & 0.93 & 0.89 & 0.91 & 0.96 \\
LR & 0.88 & 0.93 & 0.96 & 0.81 & 0.91 & 0.87 & 0.88 & 0.93 \\
NB & 0.69 & 0.93 & 0.98 & 0.40 & 0.81 & 0.56 & 0.69 & 0.87 \\
KNN & 0.85 & 0.90 & 0.94 & 0.78 & 0.89 & 0.83 & 0.86 & 0.91 \\
SVC & 0.89 & 0.95 & 0.97 & 0.83 & 0.93 & 0.89 & 0.90 & 0.94 \\
\textbf{NN} & \textbf{0.90} & \textbf{0.94} & \textbf{0.96} & \textbf{0.85} & \textbf{0.93} & \textbf{0.89} & \textbf{0.91} & \textbf{0.96} \\
\bottomrule
\end{tabular}
\vspace{-0.15in}
\end{table}

\begin{table}[t]
    \caption{Sample count per category and predicted disengaged data points.}
    \label{tab:risk_info}
    \centering
    \begin{tabular}{lllll}
        \toprule
         & high & medium & low & engaged \\
        \midrule
        No. samples & 2013 & 1578 & 84 & 4915 \\
        Predicted as disengaged & 2011 & 1234 & 6 & 3 \\
        \bottomrule
    \end{tabular}
    \vspace{-0.25in}
\end{table}

Table~\ref{tab:evaluation-metrics} assesses the performance of the implemented ML models on the test dataset. The RF model achieves a balanced accuracy of 90\% and an AUC value of 0.96. The NB and KNN models perform the worst across all metrics, indicating their inability to effectively capture the non-linearity required to differentiate between engaged and disengaged classes in the collected dataset. ML models like LR, XGB, and DT exhibit performance closer to RF in terms of accuracy and AUC metrics. However, the NN and GBM models outperform all other models, achieving an accuracy of 91\% and an AUC of 0.96. Looking to other metrics, we find that PPV and TNR (see Table~\ref{tab:metrics}) are equal, as are the F1-scores for both classes. The only differences lie in NPV and TPR (see Table~\ref{tab:metrics}), where the GBM shows a 1\% improvement compared to the NN. Nonetheless, the GBM is significantly more computationally intensive during the grid-search optimisation, training and testing compared to the NN model. In response to RQ2, the NN model is optimal to detect student disengagement from voluntary quizzes.
% \begin{figure}[t]
%     \vspace{-0.1in}
%     \centering
%     \includegraphics[width=8.5cm]{figure_4.png}
%     \vspace{-0.15in}
%     \caption{The performance of the model on the test dataset  for each day the prediction was made. }
%     \label{figure_4}
%     \vspace{-0.2in}
% \end{figure}

\vspace{-0.15in}
\subsection{RQ3: How might the explainability of ML decisions be interpreted in the context of potential interventions?}
To address this research question, we use the SHAP method to explain and interpret the decisions made by the NN model, which, as discussed earlier, is the best-performing model. We begin by analyzing the distribution of SHAP values for each feature listed in Table~\ref{tab:mean-sd-features}, identifying the most influential ones. Based on the importance ranking of these features in the decision-making process, we establish rules that reflect different engagement behaviors—such as erratic, delayed, irregular, or full engagement. Once these rules are defined across the test dataset, educators and practitioners can utilise them to interpret the NN's decisions when predicting disengagement from voluntary quizzes in Moodle. This approach provides valuable insights into student behavior, making the model's predictions more transparent and actionable.

\begin{figure}[t]
    \centering
    \includegraphics[width=9cm]{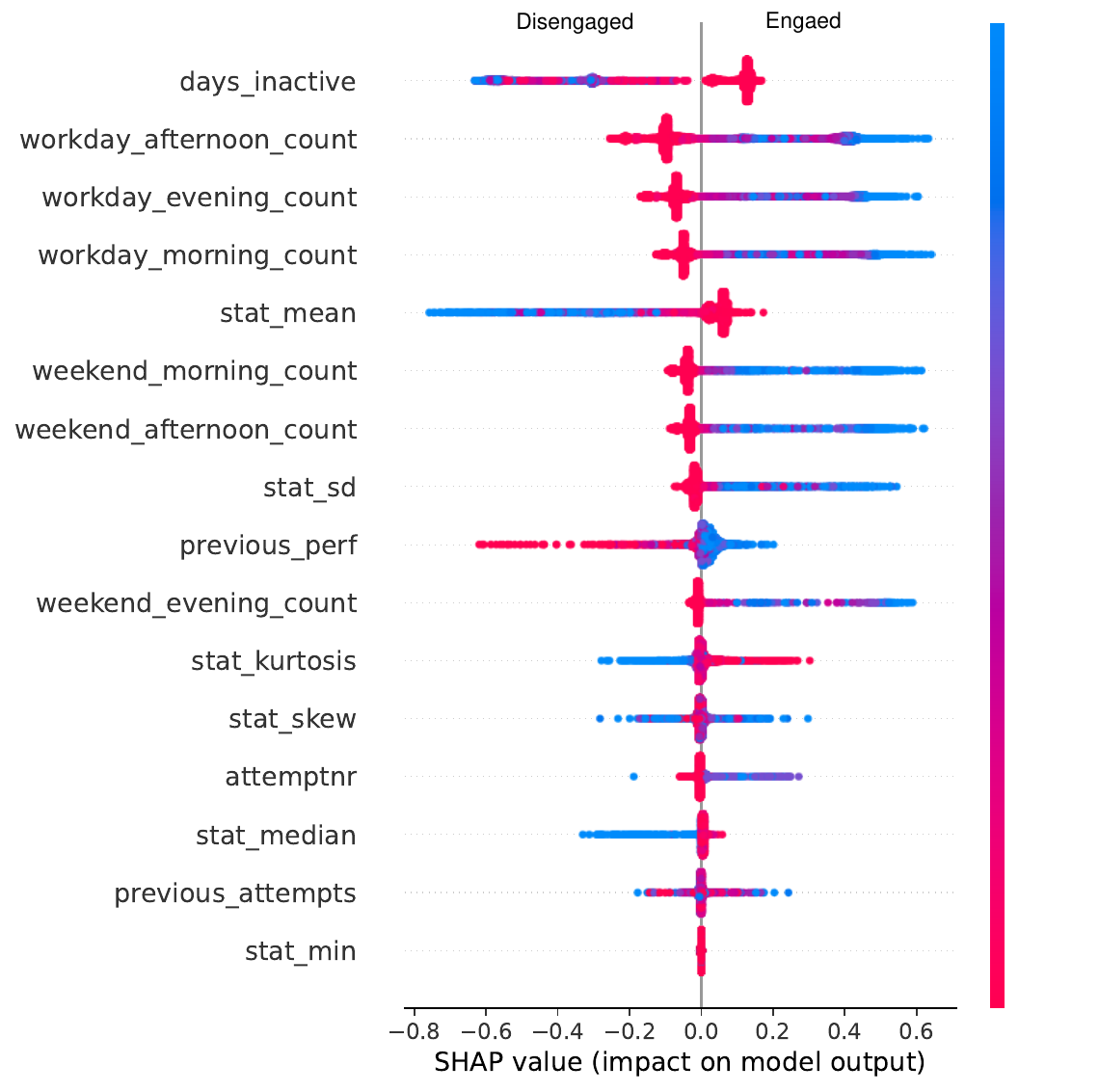}
    \vspace{-0.05in}
    \caption{SHAP values of the features for test dataset }
    \label{figure_5}
    \vspace{-0.2in}
\end{figure}

\begin{figure}[t]
    \centering
    \includegraphics[width=0.9\linewidth]{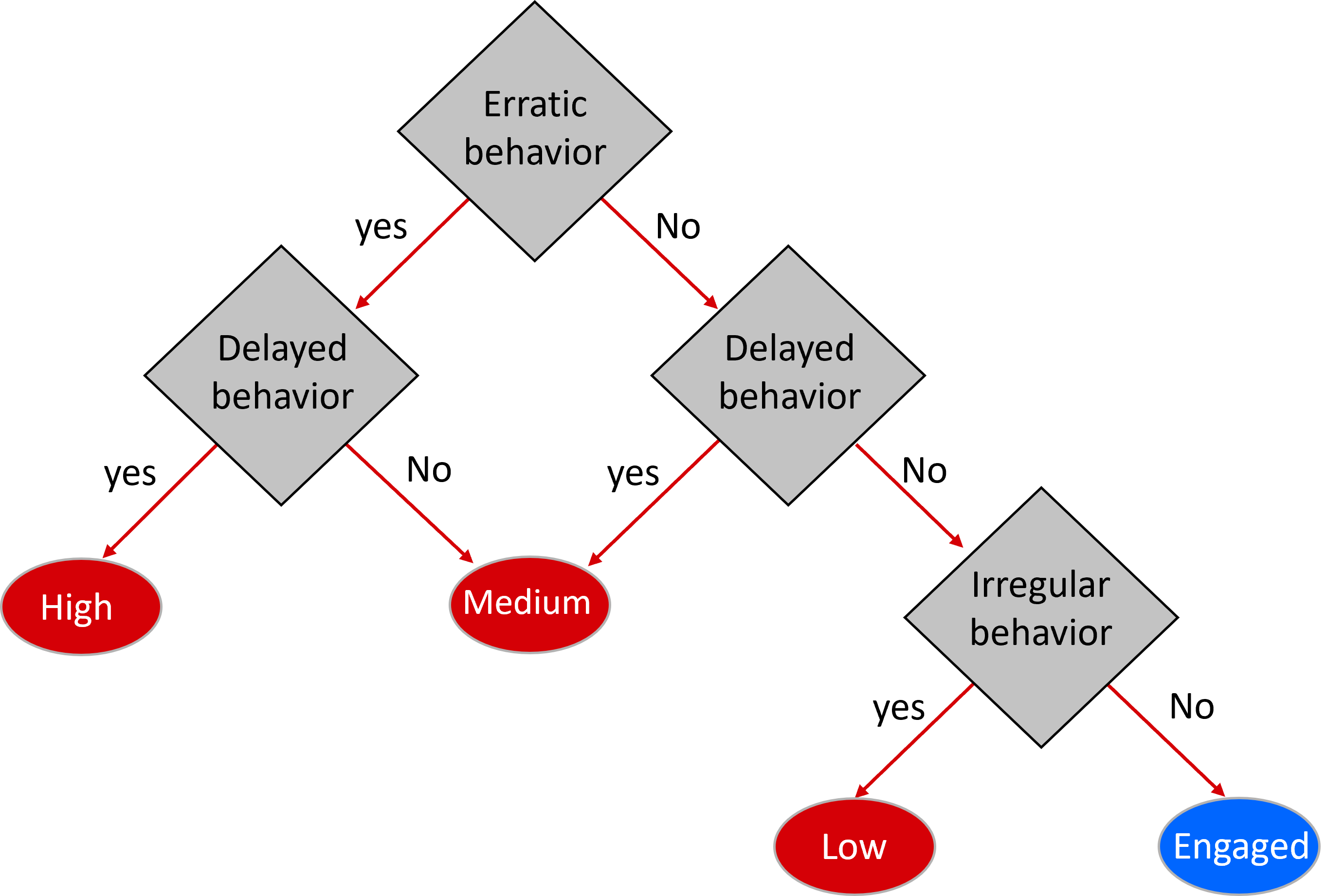}
    \caption{Determining the risk level of disengagement based on: erratic, delayed, and irregular behaviours.}
    \label{figure_6}
\vspace{-0.25in}
\end{figure}

\begin{figure*}[t]
    \centering
    \includegraphics[width=16.5cm]{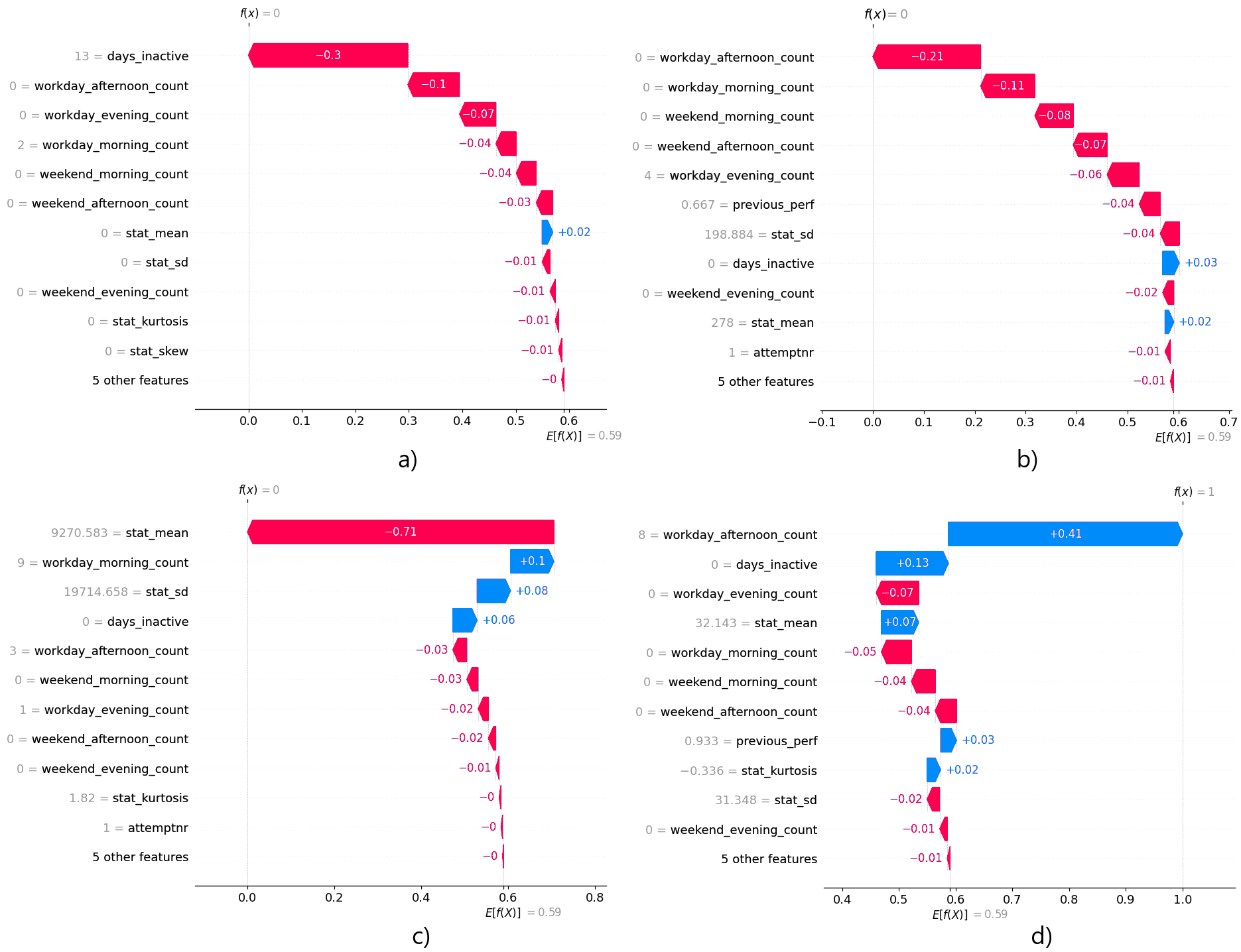}
    \vspace{-0.1in}
    \caption{SHAP Explainability Examples of a) High-risk disengagement, b) Medium-risk disengagement, c) Low-risk disengagement, and d) Engagement}
    \label{figure_7}
    \vspace{-0.2in}
\end{figure*}

Figure~\ref{figure_5} shows the distribution of SHAP values for each feature in the test dataset for the NN model. For features like, \textit{workday\_afternoon\_count}, \textit{workday\_evening\_count}, and \textit{workday\_morning\_count}, smaller values correspond to negative SHAP values which tend the model's prediction to non-submission while relatively higher values correspond to positive SHAP values which tend the model's prediction to submission. The same pattern can be observed in case of weekend days, which means that, higher number of activities (both in workday or weekend or any time of the day) leads to positive SHAP values and therefore tends the model's prediction toward submission due to higher interaction with Moodle. In contrast, for \textit{days\_inactive}, smaller values correspond to positive SHAP values or prediction of submission, while larger values indicate negative SHAP values or prediction of non-submission, suggesting that more frequent activities increases the likelihood of the model predicting submission. Similarily, for \textit{stat\_mean}, smaller values of this feature leads to positive SHAP values which tend the model's prediction toward submission while higher values of this feature lead to negative SHAP values which tend the model's prediction to non-submission. As \textit{stat\_mean} is the average of time difference between interactions of students with the quiz views, this result suggests that the model predicts submission for more frequent activities in shorter periods. At opposite, lower values of \textit{stat\_sd} are in favor of non-submission, while higher values indicates submission. In case of \textit{previous\_perf}, higher SHAP values result in predictions toward submission, while lower values lead the model to predict non-submission.

The distribution of SHAP values in Figure~\ref{figure_5} plays a crucial role in identifying the most influential features and establish rules for specific behaviors that lead to disengagement. By selecting the top 10 most influential features from this figure, we can define erratic behavior based on the number of views during weekdays and weekends, across morning, afternoon, and evening hours, as well as prior performance. Delayed behavior is linked to \textit{days\_inactive}, while the irregular behaviour is captured by \textit{stat\_mean} and \textit{stat\_sd}. Using these insights, if the sum of SHAP values for view counts is less than zero, then the erratic behavior leads to disengagement. Negative SHAP values for \textit{days\_inactive} indicate a delayed behaviour that leads to disengagement, while positive SHAP values imply higher chances to have engagement. The irregular behaviour leads to disengagement if the sum of SHAP values for \textit{stat\_mean} and \textit{stat\_sd} is negative; and the engagement is inferred, if neither of these conditions is met.

In Figure \ref{figure_6} we show how the risk levels are defined based on these bahaviours. As shown, a high risk of disengagement is detected when both erratic and delayed behaviours are present. Medium risk is associated with either erratic behaviour or with delayed behaviour but not both, while low risk is linked to the irregular behaviour only. Full engagement is identified when none of the three types of behaviours are observed.
In Table~\ref{tab:risk_info} we show the distribution of risk levels in the test dataset. For example, out of 2013 cases identified as high-risk based on SHAP values and behavioural patterns, the model predicted 2011 of them as disengaged while this ratio is 1234 out of 1578 for the medium-risk, 6 out of 84 for the low-risk and 3 out of 4915 for the engaged case. 

To better illustrate how the explainability of the SHAP method can be interpreted, and to address RQ3, we examine an example for each type of behavior. Figure~\ref{figure_7} presents the SHAP values of various features that contribute to the model’s engagement or disengagement decisions, categorised by four behavior types: erratic, delayed, irregular, and fully engaged. In Figure \ref{figure_7}(a), we present an example where erratic and delayed behaviors are identified. Delayed behavior is detected due to a negative SHAP value of \textit{inactive\_days} (caused by high values for this feature, 13), erratic behavior is identified by the negative sum of SHAP values for activity count features (workday/weekend, morning/afternoon/evening counts), and irregular behavior is not observed because of the positive sum of SHAP values for statistics-based features: \textit{stat\_mean} and \textit{stat\_sd}. According to Figure \ref{figure_6}, this example aligns with high-risk. The model’s prediction for this student was disengagement. In Figure \ref{figure_7}(b), we show an example where erratic and irregular behaviours are detected, but delayed is not observed. This is a medium-risk case, and the model similarly predicts disengagement. In Figure \ref{figure_7}(c), irregular behavior is observed, caused by a negative SHAP value for the sum of statistics-based features (\textit{stat\_mean} and \textit{stat\_sd}), while delayed and erratic behaviors are absent. This is the case of low-risk where the model predicted disengagement. Finally, Figure \ref{figure_7}(d) shows a case where none of the erratic, delayed, or irregular behaviors are present and the model predicted engagement. In Section \ref{sec:impli-from-theo-to-prac}, we identify thresholds for features and relate these risk levels to specific feature values.

{
\begin{table*}[t]
\centering
\caption{Intervention Strategies for Different Engagement Behaviors}
\label{tab:interventions}
\renewcommand{\arraystretch}{1.3} % Adjust row spacing
\begin{tabular}{|m{3cm}|m{4cm}|m{9.5cm}|}
\hline
\textbf{Behavior Type} & \textbf{Description} & \textbf{Intervention Strategies} \\ 
\hline
\textbf{Erratic Behavior} & Students displaying erratic interaction patterns, characterized by significant differences in activity between weekdays and weekends. & \begin{itemize}
    \item {\textbf{Structured Learning Plans}: Regular reminders and structured plans to establish a consistent learning routine \cite{structured_learning_plans}.}
    \item {\textbf{Gamification Elements}: Rewards (e.g., points or badges) for consistent engagement on multiple days per week \cite{gamification_elements}.}
    \item {\textbf{Increased Flexibility}: Negotiable alternative time slots or extended deadlines \cite{increased_flexibility}.}
\end{itemize} \\ 
\hline
\textbf{Delayed Behavior} & Students with long periods of inactivity, identified by high \textit{days\_inactive} values, indicating difficulty in maintaining consistent engagement. & \begin{itemize}
    % \item \textbf{Regular Check-ins}: Automated notifications to instructors, prompting personal meetings or emails.
    \item {\textbf{Motivational Messages}: Personalized messages emphasizing the importance of learning \cite{motivational_messages} (e.g., "\textit{It’s great to keep going! Every step brings you closer to your goal.}").}
    \item {\textbf{Deadline Reminders}: Regular alerts about upcoming deadlines \cite{deadline_reminders}.}
    \item {\textbf{Progressive Deadlines}: Introducing milestones to reduce overwhelm and promote regular work habits \cite{progressive_deadlines}.}
\end{itemize} \\ 
\hline
\textbf{Irregular Behavior} & Students showing inconsistent time intervals between interactions, detected through high \textit{stat\_mean} and \textit{stat\_sd} values, making engagement unpredictable. & \begin{itemize}
    \item {\textbf{Self-Reflection Feedback}: Dashboards displaying engagement patterns to improve time management \cite{self_reflection_feedback}.}
    %\item \textbf{Calendar Integration}: Synchronization with calendar apps to schedule regular learning blocks.
    \item {\textbf{Individualized Time Slots}: Suggestions for learning times based on historical activity data (e.g., morning hours for frequent interactions) \cite{individualized_time_slots}.}
    \item {\textbf{Adapted Quiz Structure}: Shorter, more frequent quizzes to encourage regular interactions \cite{adaptive_quiz_structure}.}
\end{itemize} \\ 
\hline
{\textbf{Engaged Behavior}} & {Students who consistently engage with voluntary quizzes and demonstrate regular interaction patterns without signs of withdrawal.} & \begin{itemize}
    \item {\textbf{Challenging Content}: Offer optional advanced tasks or bonus quizzes to promote deeper learning and maintain motivation \cite{challenging_content}.}
    \item {\textbf{Peer Mentorship}: Encourage engaged students to mentor peers with lower engagement, reinforcing their own commitment while supporting others \cite{peer_mentorship}.}
    \item {\textbf{Recognition of Achievement}: Use digital badges, certificates, or public acknowledgment to reward and reinforce consistent voluntary engagement \cite{reward_systems}.}
\end{itemize} \\ 
\hline
\end{tabular}
\vspace{-0.2in}
\end{table*}
}

\subsection{RQ4: How can 
%dialog-based 
intervention strategies be aligned with detected behavioral patterns to inform targeted and timely support actions?}\label{sec:designing-dialog}

{Erratic Behavior suggests a lack of routine and difficulty in maintaining consistent study habits. To mitigate this, several intervention strategies have been explored. Structured learning plans, exemplified by the Check-In/Check-Out (CICO) system, provide students with consistent routines and feedback mechanisms \cite{structured_learning_plans}. A meta-analytic review found that CICO effectively reduces disruptive behaviors and enhances academic engagement by establishing clear expectations and regular monitoring \cite{structured_learning_plans}. This structured approach can help students with erratic patterns develop consistent study habits. Gamification elements, such as points and badges, have been shown to increase motivation and engagement. Gamification positively impacts student behavior, particularly when aligned with individual personality traits \cite{gamification_elements}. By rewarding consistent participation, gamified systems can encourage regular engagement among students with irregular study habits. Lastly, Increased flexibility in learning environments addresses the diverse needs of students by offering negotiable deadlines and varied time slots. Aligning flexible learning dimensions with student and educator perspectives fosters inclusivity and accommodates varying engagement patterns \cite{increased_flexibility}. Such flexibility can help students manage their schedules more effectively, promoting steadier participation. Collectively, these strategies offer a multifaceted approach to supporting students exhibiting erratic engagement behaviors.}

{Students exhibiting delayed behavior often experience long periods of inactivity, commonly due to a lack of motivation, poor time management, or difficulty initiating tasks. To address these challenges, several evidence-based intervention strategies have been identified. Personalized motivational messages grounded in achievement goal theory significantly enhance learner engagement by reinforcing the value and relevance of academic effort \cite{motivational_messages}. These kinds of messages can provide delayed students with the psychological boost needed to re-engage with learning materials. Complementing this, in \cite{deadline_reminders} authors found that students highly value regular reminders from instructors about upcoming deadlines, noting their effectiveness in maintaining focus and accountability in online courses. Such reminders are particularly useful for students who tend to postpone tasks until the last moment. Finally, breaking down 
%large 
projects into smaller milestones significantly reduced procrastination and improved overall outcomes as shown in \cite{progressive_deadlines}. This structured pacing can be especially effective for delayed learners by lowering the cognitive and emotional barrier to getting started. Together, these interventions offer a targeted and supportive approach for helping students with delayed behavior reestablish 
%consistent 
academic engagement.}

{Students displaying irregular behavior exhibit unpredictable intervals between interactions, often resulting in unstable or fragmented engagement. To address this, interventions that promote regularity and self-awareness have shown promising results. In \cite{self_reflection_feedback}, the authors highlighted the effectiveness of self-reflection feedback through intelligent tutoring systems like MetaTutor, which use multichannel data to scaffold self-regulated learning. Dashboards that visualize engagement patterns can help students recognize irregularities in their study habits and encourage better time management. Complementing this, a modular rule-based approach for personalized recommendations is proposed in \cite{individualized_time_slots}, including scheduling suggestions based on historical engagement data. By identifying preferred or high-performing time slots (e.g., consistent morning activity), systems can encourage learners to adopt more regular study routines. Additionally, by implementing shorter, more frequent online quizzes significantly can increase student preparation and class participation \cite{adaptive_quiz_structure}. This structure reduces the cognitive load of infrequent, high-stakes tasks and promotes steady engagement through low-stakes repetition. Together, these interventions tackle the unpredictability inherent in irregular behavior by offering personalized feedback, schedule optimization, and structured task pacing, each fostering a more consistent and sustainable learning rhythm.}

{Even students not identified as disengaged from voluntary quizzes in adaptive online courses can benefit from strategies that sustain and deepen their engagement. First, offering challenging content, such as optional advanced quizzes or bonus questions, can boost motivation by supporting autonomy and promoting deeper learning \cite{challenging_content}. Second, peer mentorship, where engaged students support less active peers, not only strengthens the mentor's commitment but also fosters a collaborative learning environment \cite{peer_mentorship}. Third, recognizing consistent participation through digital badges or public acknowledgment helps sustain motivation and reinforces the value of ongoing engagement \cite{reward_systems}.}

{Table~\ref{tab:interventions} outlines the recommended interventions tailored to each type of behaviour identified. Based on the associated risk levels of disengagement from voluntary quizzes derived from these behavioral patterns, the targeted intervention planning prospects for each risk category are outlined as follows:}
\begin{itemize} 
    \item {Students identified at high risk of disengagement from voluntary quizzes exhibit both erratic and delayed behavior, 
    %characterized by inconsistent activity across weekdays and weekends, coupled with long inactivity periods before quiz completion. These learners 
    they are the most likely to skip voluntary quizzes entirely. To support them, instructors should implement structured learning plans that re-establish consistent routines, such as recommending fixed time slots for quiz work (e.g., two short sessions per week) or offering weekly planning templates integrated into Moodle \cite{structured_learning_plans}. Personalized motivational messages can be triggered after sustained inactivity, reinforcing the importance of voluntary quizzes for mastery and reinforcing a sense of academic momentum \cite{motivational_messages}. Additionally, although no strict deadlines exist, instructors can provide recommended submission windows or flexible pacing guides to help students structure their efforts and reduce procrastination-related disengagement \cite{increased_flexibility}. Together, these interventions offer a supportive structure that addresses both behavioral inconsistency and motivational inertia, guiding high-risk students back toward regular quiz participation.}

    \item {Medium-risk students exhibit either erratic or delayed engagement patterns, making them partially engaged but still vulnerable to missing voluntary quizzes. When erratic behavior dominates, gamification strategies (awarding badges or visual progress indicators for consecutive quiz submissions) can encourage routine participation and help turn irregular use into predictable engagement patterns \cite{gamification_elements}. In cases of delayed behavior, timely reminders (aligned with exam preparation periods or set intervals throughout the semester) can be sent to nudge students to attempt quizzes before content becomes stale or the final assessment nears \cite{deadline_reminders}. While voluntary quizzes have no enforced deadlines, instructors can suggest progressive timelines, such as completing one quiz per week or targeting certain quizzes before key course checkpoints. This pacing helps reduce the psychological burden of catching up and provides smaller, manageable engagement points for students prone to postponement \cite{progressive_deadlines}. These light-touch interventions aim to encourage regular participation before disengagement sets in more deeply.}

    \item {Low-risk students are generally active but interact with course materials at unpredictable times, increasing the chance of accidental or patternless non-submission of voluntary quizzes. For these students, self-reflection tools such as dashboards that visualize quiz interaction timelines can raise awareness of irregular behavior and support better planning \cite{self_reflection_feedback}. In parallel, suggesting individualized quiz prep windows, based on historical patterns of high activity (e.g., "You’re most active Mondays at 10AM"), can help form stable routines \cite{individualized_time_slots}. Structuring voluntary quizzes as smaller, more frequent activities also encourages rhythm and continuity, ensuring that even those with variable schedules stay engaged \cite{adaptive_quiz_structure}. 
    %These strategies aim to stabilize engagement and sustain voluntary quiz participation over time without overburdening already active learners.
    }
\vspace{-0.05in}
\end{itemize}
{Table~\ref{tab:risk_mapping} presents a risk-based intervention planning, where strategies are aligned with specific levels of disengagement risk, as inferred from students’ behavioral patterns.}
%\textcolor{blue}{Even students who are not identified as disengaged from voluntary quizzes in adaptive online courses can benefit from strategies designed to maintain and enhance their engagement. Firstly, offering challenging content, such as optional advanced quizzes or bonus questions, can further motivate these students by supporting autonomy and promoting deeper learning engagement. Research indicates that incorporating optional assessments can improve student perceptions of online learning and encourage deeper learning approaches\cite{challenging_content}. Secondly, encouraging peer mentorship, where consistently engaged students provide support to peers with lower engagement, not only reinforces the mentor's commitment but also contributes to a collaborative learning environment. Studies have shown that peer mentoring relationships in higher education offer roles, risks, and benefits that can positively impact student engagement\cite{peer_mentorship}. Lastly, providing recognition of consistent participation, such as digital badges or public acknowledgment, helps maintain motivation and affirms the value of sustained voluntary engagement. Evidence suggests that recognition and reward systems can enhance faculty motivation and improve student engagement and learning outcomes\cite{reward_systems}.}

{
\begin{table*}[t]
\centering
\caption{{Risk-Based Intervention Planning Informed by Detected Disengagement Behaviors}}
\label{tab:risk_mapping}
\renewcommand{\arraystretch}{1.3}
\begin{tabular}{|m{1.5cm}|m{9.5cm}|m{5.5cm}|}
\hline
{\textbf{Risk Level}} & {\textbf{Behavior Mapping and Explanation}} & {\textbf{Intervention Strategies}} \\
\hline
{\textbf{High Risk}} & {Students exhibit both erratic and delayed behaviors, marked by inconsistent weekday/weekend activity and prolonged inactivity periods. These learners are the most likely to skip voluntary quizzes entirely.} & 
{Structured Learning Plans \cite{structured_learning_plans}, Motivational Messages \cite{motivational_messages}, Increased Flexibility \cite{increased_flexibility}} \\
\hline
{\textbf{Medium Risk}} & {Students show either erratic or delayed engagement. They participate sporadically but are vulnerable to lapses. Erratic students benefit from routine incentives, while delayed students benefit from gentle nudges and pacing support.} &
{Gamification Elements \cite{gamification_elements}, Deadline Reminders \cite{deadline_reminders}, Progressive Deadlines \cite{progressive_deadlines}} \\
\hline
{\textbf{Low Risk}} & {Students are generally active but interact irregularly, risking missed quizzes due to lack of structure. Interventions aim to stabilize participation patterns.} &
{Self-Reflection Feedback \cite{self_reflection_feedback}, Individualized Time Slots \cite{individualized_time_slots}, Adapted Quiz Structure \cite{adaptive_quiz_structure}} \\
\hline
\end{tabular}
\vspace{-0.2in}
\end{table*}
}
%\textcolor{blue}{In the context of detecting disengagement from voluntary quizzes in adaptive online courses, the timing of interventions should be aligned with both the severity of the risk and the interpretability of the detection results. For students classified as high risk, intervention should occur immediately after the first signs of disengagement are detected, as this group shows the highest likelihood of completely skipping voluntary quizzes. Given that these patterns are strongly associated with clear thresholds in observable features (e.g., prolonged inactivity or lack of weekday engagement), instructors can act promptly without relying on deep model interpretability. In contrast, for medium- and low-risk students, whose disengagement stems from more nuanced patterns like irregular or isolated erratic behavior, the timing of interventions should be informed by SHAP-based explainability outputs. Here, instructors may wait until the model identifies emerging negative SHAP contributions across successive activities, signaling a downward engagement trend. Interventions for these students can be delivered periodically, such as after each quiz window or at predefined course checkpoints, when model outputs indicate a drift toward disengagement, but before behavioral patterns become entrenched. This tiered approach ensures that interventions are timely, proportionate to risk, and grounded in actionable analytics.}
{When detecting disengagement from voluntary quizzes in adaptive online courses, the timing of interventions should match both the risk level and the clarity of detection. For high-risk students, immediate intervention is essential, as they are most likely to skip quizzes entirely. Since their behavior aligns with clear thresholds (e.g., prolonged inactivity or low weekday engagement), instructors can act without relying on model interpretability. For medium- and low-risk students, whose disengagement stems from subtler patterns, SHAP explainability should guide timing. Instructors can intervene when negative SHAP values accumulate across activities, indicating a downward trend. These interventions can occur after quiz windows or at course checkpoints, before habits worsen. This tiered strategy ensures interventions are timely, risk-sensitive, and grounded in actionable insights.}

\vspace{-0.1in}
\section{Discussion} \label{sec:discussion}
{This section begins by positioning our proposed approach in relation to existing state-of-the-art methods, demonstrating its comparative strengths and relevance in the context of distance and adaptive learning environments. We then examine the practical implications of our findings, particularly how explainable disengagement detection can inform targeted instructional strategies. Finally, we reflect on the current limitations of the framework and propose concrete directions for future research and development, including the implementation and evaluation of 
%dialog-based 
interventions.}
%We present the limitations of the proposed framework, compare its performance with other state-of-the-art approaches, and explore the potential impact of this solution in the field of distance education. We also propose various dialogue-based intervention strategies tailored to each identified risk category of disengagement from voluntary tasks.

\vspace{-0.15in}
\subsection{Performance Comparison with State-of-the-art}
Our framework, achieving a maximum accuracy of 91\% with NN and GBM models, is compared with related state-of-the-art approaches, though most do not specifically target disengagement in voluntary tasks. Cocea et al. \cite{cocea_2009} achieved 95\% accuracy with a LR model for engagement prediction from logs and quiz results. Hussain et al. \cite{hussain_2018} reached 89\% accuracy using a J48 decision tree based on activity types and clicks, while Raj et al. \cite{raj_2022} obtained 94\% accuracy with RF model including final exam scores. In reading tasks, Mills et al. \cite{mills_2014} reported 89\% accuracy for predicting when students would quit after reading the first page. Despite focusing on voluntary quiz interactions, our model demonstrates competitive performance and, through SHAP-based explainability, accurately identifies disengagement categories, enabling targeted interventions for each risk group.

\begin{figure*}[t]
    \centering
    \includegraphics[width=18cm]{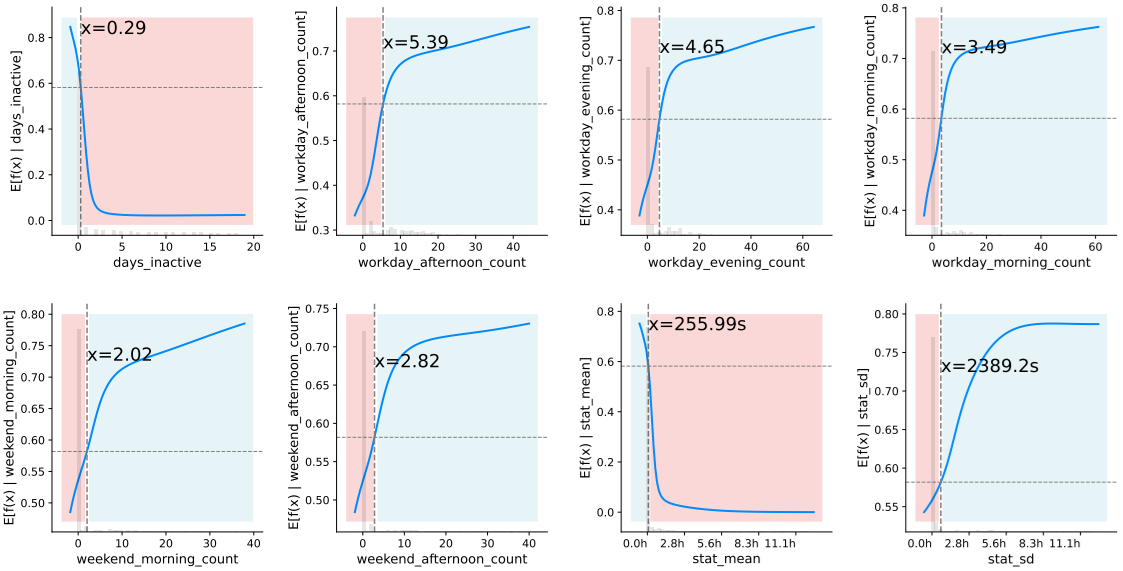}
    \caption{Partial dependence plots showing the relationship between the top eight most important features and the model's predicted probability of engagement. The horizontal dashed line represents the models' expected probability while the vertical dashed lines with labels indicate the feature thresholds where predictions shift significantly. Red/blue regions show the negative/positive SHAP values region respectively. }
    \label{figure_8}
    \vspace{-0.2in}
\end{figure*}

\vspace{-0.15in}
\subsection{Implications: From Theory To Practice}\label{sec:impli-from-theo-to-prac}
In their work on the theoretical foundation of engagement in online learning, Saqr et al. \cite{saqr_trajectory_2021} categorise students into three groups: actively engaged, averagely engaged, and disengaged. Active engagement is characterised by factors such as the number of active days on the platform, course browsing frequency, forum participation, regularity, and session count. For the averagely engaged group, key contributing factors include the frequency of forum consumption, forum contributions, and lecture viewing. In contrast, disengaged students are primarily influenced by low levels of forum consumption and contribution. This framework can be extended to detect disengagement in voluntary tasks, with the disengaged group further divided into high, medium, and low-risk categories. These levels closely align with the behaviors illustrated in Figure~\ref{figure_6}. For example, a high risk of disengagement is linked to infrequent quiz views, prolonged inactivity, and low forum participation. To more accurately identify students at medium or low risk of disengagement, it is important to include specific forum activity features in the analytics and explainability framework. As highlighted by Saqr et al., forum indicators can help identify the disengaged class in general, which can then be further categorised based on the behaviors from the voluntary tasks as outlined in Fig.~\ref{figure_6}.

%\vspace{-0.1in}
This paper emphasises the practical implications of detecting behavioural disengagement in online and adaptive learning, focusing on how educators can utilise the proposed framework. Beyond identifying disengagement risk through student interaction data, the framework explains decisions clearly, enabling educators to design effective interventions. Figure~\ref{figure_8} presents SHAP dependence plots, which depict the relationship between specific features and the model's predicted engagement probability (a probability of 0 and 1 means disengagement and engagement respectively). Each plot shows the expected value of the model's prediction, $E[f(x)\mid feature]$, given the value of a particular feature, while averaging over the effects of all other features. These plots highlight how variations in a feature (e.g., days inactive or workday counts) influence the model’s prediction. The dependence plots shown correspond to the top 8 most important features identified in Figure 4, providing deeper insights into their contribution to the model's behavior. We also provide practical thresholds for identifying disengagement (red regions) or engagement (blue regions) in the figure. For instance, Figure~\ref{figure_7}(a) and cross-referencing it with Figure~\ref{figure_8}, we observe 13 inactive days which are well above the threshold of 0.29 days. Additionally, all erratic-related features fall below their respective thresholds, making it easy to classify this observation as high risk for disengagement. Similarly, in Figure~\ref{figure_7}(b), inactivity is below the threshold, but all erratic features remain under the red region shown in Figure~\ref{figure_8}, which points to a medium risk of disengagement. These thresholds enable timely feedback for high- and medium-risk cases. For low-risk or fully engaged students, where feature contradictions often arise, SHAP’s explainability is essential for quantifying feature contributions and ensuring accurate classification between disengagement and engagement.

% \vspace{-0.1in}
The rules derived from trained ML algorithms and SHAP-based explainability can be seamlessly integrated into LMS platforms (e.g., Moodle) to enhance learning experiences by enabling: \textit{a)} dynamic adjustments, allowing automatic modifications to quiz layouts or difficulty levels based on detected disengagement levels; and \textit{b)} personalized recommendations, providing tailored activity suggestions aligned with individual behavioral patterns to foster higher engagement. Leveraging SHAP for model interpretation promotes transparency and acceptance by benefiting both: \textit{a)} instructors, who gain valuable insights into why certain students are classified as disengaged, facilitating the creation of targeted intervention strategies; and \textit{b)} students, who receive clear and transparent feedback, empowering them to understand and independently improve their learning behaviors.

{From a practical standpoint, implementing the proposed %dialog-based 
interventions requires a structured integration with the LMS, where each intervention strategy is triggered based on the student's predicted disengagement risk level. The following steps outline how this can be operationalized: 
%as part of our future work:
}

\begin{enumerate}
    \item {\textbf{Data Collection and Feature Extraction:} Moodle should continuously preprocess student interaction data (e.g., quiz activity timestamps, session frequency, inactivity durations, etc.). These features feed into the disengagement detection pipeline.}

    \item {\textbf{Model Inference and Risk Classification:} A trained ML model (e.g., a neural network) is deployed to classify students based on real-time data. Using SHAP explainability, predictions are interpreted to assign a disengagement risk level, based on specific combinations of erratic, delayed, or irregular behaviors.}

    \item {\textbf{Mapping Risk Levels to Interventions:} High-risk students require immediate and structured support, combining motivational nudges with routine-building strategies and flexible pacing guidance. Medium-risk students benefit from periodic prompts, such as gamified engagement incentives or progressive activity reminders to prevent further disengagement. Low-risk students are supported through stability-focused interventions, including visual self-monitoring tools, personalized scheduling suggestions, and frequent low-stakes quiz formats. Finally, engaged students should receive interventions aimed at sustaining motivation, including optional challenges, peer mentoring opportunities, and recognition of continued participation.}

    \item {\textbf{Automated Delivery and Instructor Oversight:} Moodle should support a lightweight intervention engine that sends messages based on the detected risk level. Instructors must retain the ability to review, personalize, or override interventions via dashboards.}

    \item {\textbf{Logging and Feedback Loops:} The system should log student reactions and behavioral changes post-intervention. This enables iterative refinement of both the predictive model and the intervention logic, supporting continuous improvement and adaptation over time.}
\end{enumerate}

{This layered implementation framework ensures that disengagement detection results in timely, personalized, and pedagogically informed 
%dialog-based 
interventions, tailored to risk level and grounded in interpretable analytics.}

\vspace{-0.15in}
\subsection{Limitations and Future Work}
{The proposed solution accurately detects disengagement from voluntary quizzes by analyzing interaction and quiz activity data within Moodle, classifying students as engaged or disengaged with 91\% accuracy. It identifies behavioral patterns, explains engagement decisions, and categorizes students into three risk levels, each linked to targeted intervention strategies. However, these interventions (though grounded in literature) have not yet been tested in live learning settings, leaving their practical effectiveness unverified. Additionally, the interpretability of the model is 
%closely 
tied to the current neural network architecture and feature engineering approach. Future work, such as integrating contextual or forum-based data, may enhance classification performance but could impact the transparency and consistency of the behavioral risk categorization.}

{Despite its limitations, this work provides a technically robust and coherent foundation. The key contributions are:}
\begin{itemize}
\item {\textbf{Comprehensive data collection:} Log data from four semesters, covering 565 students and over 30,000 interaction records.}
\item {\textbf{Targeted feature engineering:} Custom features derived exclusively from voluntary quiz activity, using semester-end as a proxy deadline.}
\item {\textbf{Model benchmarking:} Evaluation of nine ML models, with a neural network selected for optimal performance.}
\item {\textbf{Behavioral explainability:} SHAP analysis revealed consistent feature patterns, enabling definition of three behavior types.}
\item {\textbf{Risk classification:} Rule-based mapping of behaviors into high, medium, and low disengagement risk levels.}
\item {\textbf{Strategy alignment:} Literature-informed interventions matched to each risk level for actionable support.}
\item {\textbf{Instructor interpretability:} SHAP-derived thresholds support instructor awareness without deep knowledge on machine learning and data science.}
\item {\textbf{Deployment readiness:} A technical roadmap for integrating detection and intervention in Moodle.}
%real-world platforms.}
\end{itemize}
%\textcolor{blue}{Future work will focus on transforming this diagnostic framework into a fully operational predictive and prescriptive learning analytics system. This includes both predictive modeling of future disengagement and implementation of the dialog-based interventions defined in this work. The following didactic steps are planned for the next phase:}
%\begin{enumerate}
%    \item \textcolor{blue}{\textbf{Integration with LMS:} embedding the disengagement detection pipeline and intervention triggers into Moodle for real-time use.}
%    \item \textcolor{blue}{\textbf{Controlled Intervention Deployment:} rolling out interventions in live courses using pilot groups, triggering actions based on risk levels and monitoring changes in engagement.}
%    \item \textcolor{blue}{\textbf{Outcome Measurement:} evaluation the impact of interventions through follow-up analysis of quiz submission rates, behavioral shifts, and learning outcomes.}
%    \item \textcolor{blue}{\textbf{Instructor Involvement:} equipping instructors with dashboards to visualize SHAP-based engagement insights, intervene manually, or override recommendations.}
%    \item \textcolor{blue}{\textbf{Iterative Refinement:} Using the outcomes and user feedback to update the detection model, refine feature thresholds, and improve the mapping between risk and intervention.}
%\end{enumerate}
%\textcolor{blue}{These future directions will move the proposed framework from a robust diagnostic solution to a dynamic, feedback-driven support system for fostering engagement in adaptive online learning environments.}
{Future work aims to evolve the current diagnostic framework into a fully predictive and prescriptive learning analytics system. This includes forecasting future disengagement and deploying the defined interventions within real course settings. The next phase involves:}
\begin{enumerate}
\item {\textbf{LMS Integration:} embedding the detection and intervention mechanisms into Moodle for real-time operation.}
\item {\textbf{Pilot Deployment:} implementing interventions in live courses, triggered by risk levels and monitored for engagement changes.}
\item {\textbf{Impact Evaluation:} assessing intervention effects via quiz submissions, behavioral trends, and learning outcomes.}
\item {\textbf{Instructor Tools:} providing dashboards for SHAP-based insights, manual intervention, and override options.}
\item {\textbf{Model Refinement:} updating detection models and intervention logic based on outcomes and instructor feedback.}
\end{enumerate}
{These steps will establish an adaptive, feedback-driven framework designed to sustain and enhance student engagement in online learning environments.}

\vspace{-0.1in}
%\section{Conclusion} \label{sec:conclusion}
%In this paper, we analyzed the behavioural disengagement from voluntary tasks in adaptive and online learning environments. We developed a diagnostic learning analytics framework capable of detecting disengagement with 91\% prediction accuracy, while also identifying different levels of disengagement risk to support the creation of tailored, dialogue-based interventions. Our findings show that erratic behaviour is tied to the number of quiz views during various time periods across weekdays and weekends. Delayed behaviour is linked to the number of inactive days since the quiz started, and irregular behaviour is connected to the inconsistency of students' interactions with different quiz activities. Using SHAP explanations, we derived insightful interpretations that associate these behaviours with varying disengagement risk levels. Specifically, high risk is linked to both erratic and delayed behaviours, medium risk is tied to either erratic or delayed behaviour, and low risk is associated with irregular behaviour. The strong predictive accuracy and clear interpretability of our framework will help instructors better understand the predictions and implement targeted interventions based on the specific disengagement risk.
\section{Conclusion} \label{sec:conclusion}
In this paper, we analyzed the behavioural disengagement from voluntary tasks in adaptive and online learning environments. We developed a diagnostic learning analytics framework capable of detecting disengagement with 91\% prediction accuracy, while also identifying different levels of disengagement risk to support the creation of tailored, dialogue-based interventions. Our findings show that erratic behaviour is tied to the number of quiz views during various time periods across weekdays and weekends. Delayed behaviour is linked to the number of inactive days since the quiz started, and irregular behaviour is connected to the inconsistency of students' interactions with different quiz activities. Using SHAP explanations, we derived insightful interpretations that associate these behaviours with varying disengagement risk levels. Specifically, high risk is linked to both erratic and delayed behaviours, medium risk is tied to either erratic or delayed behaviour, and low risk is associated with irregular behaviour. The strong predictive accuracy and clear interpretability of our framework will help instructors better understand the predictions and implement targeted interventions based on the specific disengagement risk. {To address each risk level, we proposed 
%dialog-based 
intervention strategies such as structured learning plans, motivational prompts, gamification elements, and self-reflection tools. Future work will focus on integrating these interventions into the learning platform, piloting them in live courses, and evaluating their impact on student engagement and learning outcomes.}

% or


% \appendices

% \section{}

%\ifCLASSOPTIONcompsoc
%\else
  %\section*{Acknowledgment}
%\fi


\ifCLASSOPTIONcaptionsoff
  \newpage
\fi
\vspace{-0.1in}
\bibliographystyle{IEEEtran}
\bibliography{ReferenceList}

\end{document}